# Towards Quantification of Bias in Machine Learning for Healthcare: A Case Study of Renal Failure Prediction

**Josie V. Williams**
jvw242@nyu.edu
Departments of Computer Science
Courant Institute of Mathematical Sciences
New York University

**Narges Razavian**
narges.razavian@nyumc.org
Departments of Population Health and Radiology
Center for Data Science
New York University Langone Medical Center

## Abstract

As machine learning (ML) models, trained on real-world datasets, become common practice, it is critical to measure and quantify their potential biases. In this paper, we focus on renal failure and compare a commonly used traditional risk score, Tangri, with a more powerful machine learning model, which has access to a larger variable set and trained on 1.6 million patients' EHR data. We will compare and discuss the generalization and applicability of these two models, in an attempt to quantify biases of status quo clinical practice, compared to ML-driven models.

## 1 Introduction

Data-driven models have become more common in the U.S. healthcare field as their use in clinical operations and diagnosing procedures have expanded exponentially. The ever-increasing processing power of machine-learning algorithms allows automatic analysis of huge quantities of data, theoretically maximizing the efficiency and accuracy of the medical diagnosing process. Predictions from machine-learning models already drive important healthcare decisions for over 70 million people across the United States[7]. However, there is reason to be cautious about assuming that the generalization of these models can be applied universally to all populations. Data-driven algorithms used in the criminal sentencing process, for example, have already shown evidence of negatively and disproportionately targeting defendants from low-income black communities due to biases inherent to the data used to train these predictive models[1]. Considering the fact that health care delivery already varies considerably by race, it is plausible to suspect that similar inequalities might be reflected in the decisions of machine-learning algorithms trained on historical data that lacks awareness and insight into certain populations.

The scope of this paper will be centered around understanding the accuracy, applicability, and fairness of a simple prediction model, Tangri[10], commonly used to predict imminent risk of kidney failure. We compare this model, with another model developed using a larger population and more comprehensive variable sets derived from Electronic Health Records (EHR), and discuss differences in performance, applicability, and generalization. This understanding will help clinical safety professionals critically appraise predictive kidney disease models from a quality perspective.

In particular, we focus on understanding whether each of the two predictive models are accurate across all populations. If not, what populations are being overlooked by each model? If we start automatic screening and interventions based on features needed to compute Tangri score, which populations will be excluded? How does a new model trained with more comprehensive variables compare to the traditional models commonly used in today's practice? Does more sophisticated machine learning make healthcare operations more or less biased?

## 2 Background and Related Works

The Tangri baseline model [10] is a simple logistic regression model, designed to predict the risk of progression towards kidney failure for patients with CKD stages 3 to 5. It has been widely adapted for the US and non-US population [9]. A basic Tangri score relies on age, gender, estimated glomerular filtration rate (eGFR) and urine Albumin to Creatinine ratio (ACR) tests to compute the risk of kidney failure. This is currently the leading model that provides hospitals and other health care providers with a method of generating a risk assessment for kidney failure and perform early interventions [4]. However, it is unclear if large scale automatic screening methods based on this model are applicable to all high-risk patients, and/or whether certain sub-populations are excluded from the benefits of screening either due to lack of available eGFR or ACR tests, or due to low predictive performance of the model. This presents the feasible possibility of overlooking potentially high-risk patients and missing crucial opportunities for early intervention.

However, machine learning models using Electronic Health Records (EHR), have recently emerged as an alternative to traditional risk prediction modeling, and provide more expressive and sophisticated methodologies for predicting future risk of kidney failure [11, 8, 3]. These models can simultaneously find surrogate variables when either of eGFR or ACR is missing, and combine all predictive variables to compute the probability of progression to kidney failure. While more expressive models have been criticized for their lack of generalization and performance on minority populations, there are very few studies that expand upon the analysis and comparison of industry standard models and models that are more advanced or accurate.

Another closely related work, Framingham[6], addressed concerns about a commonly used risk score for cardiovascular events. It showed that the presence of atherosclerotic disease interacted differently between race and ethnic groups[2]. They suggested that factors such as age or blood pressure should be weighed differently in order to retain predictive performance for different racial and ethnic backgrounds. We explore this topic in the context of kidney failure.

## 3 Methods

### 3.1 Development Cohort

The development cohort was compiled from the electronic health records (EHR) of 1.6 million patients who received care at NYU Langone Medical Center. For each patient, we used all demographics, disease history, lab results, and procedures time series between 2014 to 2018, and focused on the set of patients who do not have kidney failure yet at the time of prediction.

Kidney failure was defined as observing at least two instances of any of the following ICD-10 codes (I953, R880, T81502A, T81502D, T81502S, T81512A, T81512D, T81512S, T81522A, T81522D, T81522S, T81532A, T81532D, T81532S, T81592A, T81592D, T81592S, T8241XA, T8241XD, T8241XS, T8242XA, T8242XD, T8242XS, T8243XA, T8243XD, T8243XS, T8249XA, T8249XD, T8249XS, T85611A, T85611D, T85611S, T85621A, T85621D, T85621S, T85631A, T85631D, T85631S, T85691A, T85691D, T85691S, T8571XA, T8571XD, T8571XS, Y622, Y841, Z4901, Z4902, Z4931, Z4932, Z9115, Z940, Z992), or dialysis or kidney transplant procedure codes CPTs (G0257, 36902, 36903, 36904, 36905, 36906, 90940, 99512, 90918, 90945). Our analysis was limited to patients who did not have kidney failure by 2016/4/1. The task was to evaluate, based on patient data between 2015/1/1 to 2016/1/1, which patients will proceed to have kidney failure between 2016/4/1 to 2017/4/1.

### 3.2 Training and Validation Cohort

We included 183,490 patients in training set for our model, and 324,685 patients were included in our validation set. Of these patients without renal failure, 98 patients in the training set, and 166 patients in the validation set proceeded to have renal failure between 2016/4/1 to 2017/4/1.

### 3.3 Variables for building model from EHR

For each patient, we used a total of 28,561 variables, derived from demographics, disease diagnosis, procedures and lab values observed between 2015/1/1 to 2016/1/1. We aggregated each patient's one year of EHR data by taking the total count of each disease and procedure observations, and maximum of each lab result observed in that year. We then binned all the non-binary data including age. Lab value bins were defined according to the standard deviation of the normalized value (if observed). Age was binned into 10 year spans. Diseases and procedures were binned into >0, >2, and >10 observations. After this pre-processing step, we only kept features that were observed in at least 10 patients in the training set (of 183,490 patients). This, lead to 28561 total variables.

### 3.4 Machine Learning Model

To build our model, We used L1-regularized logistic regression model[5] to train a prediction model on the training set. We tuned the regularization parameter between values of 0.0001, 0.001, 0.002, 0.005, 0.007, and 0.01, and report results on our validation set. We used weighted log loss, to deal with our unbalanced data.

### 3.5 Computation of Tangri Scores

Tangri score in its simplest form requires age, gender, eGFR and Albumin to Creatinine ratio(ACR). For eGFR, we used any of the following labs we had observed between 2015/1/1 to 2016/1/1 : *Glomerular filtration rate/1.73 sq M.predicted [Volume Rate/Area] in Serum or Plasma (MDRD or Schwartz method)* which included any of LOINC codes 33914-3, 48642-3, 48643-1, 50210-4, 50384-7, 62238-1, 69405-9.

For ACR, we used any of the lab measurements corresponding to *Albumin/Creatinine [Ratio] in Urine*, LOINC codes 32294-1, 14585-4, 30000-4, 30001-2, 32294-1, 59159-4, 76401-9, 77253-3, 77254-1, 9318-7, 44292-1, 14959-1, 14958-3, 13705-9, 9318-7, 58447-4. The unit of these labs are taken into account (either mg/L or molar). If we have multiple labs for either eGFR or ACR, as long as the labs were in comparable unit, we take the maximum value of all of them.

Once we selected the age, gender, eGFR and ACR, we used the following formula to compute the probability of kidney failure according to original 4-variable Tangri score[10]:

$$P_{Tangri}(RenalFailure) = 1 - 0.9751^S$$

where $S = exp(-0.2201 \times (age/10 - 7.036) + 0.2467 \times (male - 0.5642) - 0.5567 \times (eGFR/5 - 7.222) + 0.4510 \times (logACR - 5.137))$

If a patient does not have eGFR or ACR, they will not get a Tangri score. This is the basis for our analysis to understand how many patients qualify for measuring Tangri, and how many patients (even if high-risk) will not be able to have a Tangri score.

## 4 Results

Table 1 includes the statistics of the total number of patients under various sub-populations, who qualified or did not qualify for Tangri, as well as how many of them proceeded to have renal failure. All the patients qualified for our model. Results are reported on the validation set.

| Cohort | Patients (had RF) | Tangri Eligibles (had RF) | Tangri-Eligible Percentage | Tangri non-Elig (had RF) |
|---|---|---|---|---|
| Full validation set | 324685 (166) | 7681 (8) | 2.4% | 317004 (158) |
| Gender: Female | 198362 (64) | 4116 (2) | 2.1% | 194246 (62) |
| Gender: Male | 126305 (102) | 3565 (6) | 2.8% | 122740 (96) |
| Race: African American | 30348 (33) | 1220 (2) | 4.0% | 29128 (31) |
| Race: Asian | 4621 (8) | 118 (0) | 2.6% | 4503 (8) |
| Race: White | 199665 (71) | 4103 (3) | 2.1% | 195562 (68) |
| Age: 20-30 | 29220 (3) | 164 (0) | 0.6% | 29056 (3) |
| Age: 30-40 | 39068 (13) | 444 (0) | 1.1% | 38624 (13) |
| Age: 40-50 | 44836 (20) | 956 (0) | 2.1% | 43880 (20) |
| Age: 50-60 | 57845 (27) | 1805 (0) | 3.1% | 56040 (27) |
| Age: 60-70 | 57780 (57) | 2150 (5) | 3.7% | 55630 (52) |
| Age: 70-80 | 36951 (26) | 1473 (2) | 4.0% | 35478 (24) |
| Age: 80-90 | 17000 (18) | 603 (1) | 3.5% | 16397 (17) |

Table 1: Patients with or without Renal Failure (RF) who did or did not qualified for Tangri scoring

We note that overall, significantly more patients who did progress to Renal Failure, did not have the lab measurements needed (eGFR and ACR) to allow them to be scored by Tangri method. This could be due to them having the tests done elsewhere, or not being managed by a nephrologist. All high-risk patients under the age of 60 in our cohort would have been missed if Tangri scoring were to be applied to EHR data retrospectively and automatically. In Table 2, we show the performance of Tangri and our model for each sub-population. We observe that our model has above a 0.84 AUC in all sub-populations, except the younger cohort (age < 40) which both Tangri and our model performed poorly on. We also observe that both models severely under-perform for the subset of female patients who had eGFR and ACR, although for the broader female cohort, including patients who did not participate in these tests, we see that our model achieves a 0.89 AUC. In addition, the Tangri model under-performs for the African American sub cohort, whereas our proposed model achieves an AUC of 0.787 on African American patients who had eGFR and ACR, and 0.885 for the general African American cohort.

| Cohort | AUC of M | AUC of Tangri on Tangri-eligible patients | AUC of M on Tangri-eligible patients | AUC of M on Tangri-non -eligible patients |
| --- | --- | --- | --- | --- |
| Full validation set | 0.88 | 0.86 | 0.836 | 0.886 |
| Gender: Female | 0.877 | 0.48 | 0.55 | 0.893 |
| Gender: Male | 0.871 | 0.992 | 0.929 | 0.872 |
| Race: African American | 0.885 | 0.554 | 0.787 | 0.909 |
| Race: Asian | 0.876 | N/A | N/A | 0.874 |
| Race: White | 0.848 | 0.995 | 0.939 | 0.848 |
| Age: 20-30 | 0.55 | N/A | N/A | 0.549 |
| Age: 30-40 | 0.788 | N/A | N/A | 0.787 |
| Age: 40-50 | 0.922 | N/A | N/A | 0.92 |
| Age: 50-60 | 0.944 | N/A | N/A | 0.943 |
| Age: 60-70 | 0.846 | 0.756 | 0.719 | 0.876 |
| Age: 70-80 | 0.853 | 0.993 | 0.968 | 0.842 |
| Age: 80-90 | 0.891 | 0.992 | 0.997 | 0.883 |

Table 2: Performance of our proposed model(M) in predicting Renal Failure compared with Tangri model on different sub-populations. N/A indicates that there were not enough Tangri-eligible patients to evaluate the AUC. All AUCs are measured on the validation set.

## 5  Discussions and Conclusion

Renal failure is devastating. Treatment options for renal failure include dialysis, a procedure that is required indefinitely three times a week, or kidney transplant, which is extremely invasive. Preventing renal failure brings years of quality life to patients and early intervention often increases the probability of successful treatments. Currently, to screen for high-risk patients, Tangri, which needs eGFR (a blood test) and ACR (a Urine test) is recommended to predict the risk of renal failure within 2 years. As we saw in Result section Table 1, these tests are not available for over 95% of patients who truly progressed to Renal failure. In a general patient cohort, the percentage of patients who already had these tests measured were even lower: 2.3%. Further still, we saw that this ratio becomes even lower in younger populations like women and several other sub-cohorts. If we were to screen all patients based on the traditional Tangri score, the measurement of eGFR and ACR alone would incur significant financial and resource costs. On the other hand we showed that by using readily available electronic health records instead of two very specific, but often unavailable tests, we can scale screening to all populations. As we saw in Table 2, the performance of the new model in predicting renal failure remains high for sub-cohorts that the Tangri risk score proved to be very precise (i.e Male, white, old). However, our model increases the performance of risk prediction for rest of the sub-cohorts (women, African American, Asian, and younger patients). Several available techniques can further improve the quality and fairness of the model across all sub-cohorts, which is part of our ongoing and future work.

In conclusion, in this paper, we focused on a well-established regression model (Tangri) and compared it with a modernized and larger model trained on EHR data. Our results indicated that (1) A model such as Tangri is already heavily biased and should be significantly scrutinized for large-scale screening due to its lack of inclusively, and (2) Machine learning models trained on a diverse cohort with access to more variables improve prediction performance for all sub-populations. This preliminary analysis is a step towards understanding and quantifying bias of status quo prediction methods, compared to stronger machine learning models.